\documentclass[conference,10pt]{IEEEtran}
\usepackage[utf8]{inputenc}

\usepackage{amsmath,amsfonts,amsthm}
\usepackage{graphicx}
\usepackage{enumitem}
\usepackage[table]{xcolor}
\usepackage{xspace}
\usepackage{tikz}
\usepackage{subfig}
\usepackage{wrapfig}
\usepackage{multirow}
\usepackage{booktabs}
\usepackage[most]{tcolorbox}
\usepackage{tabularx,collcell}
\usepackage[numbers,compress]{natbib}

\newcommand{\egospec}{{\psi_\ego}}

\newcommand{\advrules}{{\varphi_\ado}}
\newcommand{\agent}{\mathcal{H}}
\newcommand{\ado}{\mathsf{ado}}
\newcommand{\ego}{\mathsf{ego}}

\newcommand{\system}{\mathcal{S}}

\newcommand{\astates}{X}
\newcommand{\initstates}{{X_\mathrm{init}}}

\newcommand{\actions}{A}
\newcommand{\transitions}{T}
\newcommand{\policy}{\pi}

\newcommand{\astate}{\mathbf{x}}
\newcommand{\action}{a}
\newcommand{\reward}{R}

\newcommand{\traj}{s}
\newcommand{\trajsig}[1]{\traj(#1)}

\newcommand{\norm}[1]{\left|\left|#1\right|\right|}
\newcommand{\absval}[1]{\left|#1\right|}
\newcommand{\discountfactor}{\gamma}
\newcommand{\rob}{\rho}

\newcommand{\setof}[1]{\{#1\}}
\newcommand{\reals}{\mathbb{R}}
\newcommand{\alw}{\mathbf{G}}
\newcommand{\ev}{\mathbf{F}}
\newcommand{\until}{\mathbf{U}}

\newcommand{\mypara}[1]{\vspace{0.3em} \noindent {\bf #1}.}
\newcommand{\myipara}[1]{\vspace{0.3em} \noindent {\em #1}.}
\DeclareMathOperator*{\argmax}{arg\,max}

\newcommand{\valuefn}{V}

\newcommand{\optimalvalfcn}[1]{\valuefn^{\star}(#1)}
\newcommand{\suboptimalvalfcn}[1]{\valuefn_{k}(#1)}
\newcommand{\expectedvalue}{\mathbb{E}}

\newcommand{\ignore}[1]{}

\newcommand{\dman}{d_\mathrm{man}}
\newcommand{\policies}{\Pi}

\newcommand{\trn}{\mathrm{train}}
\newcommand{\egoinit}{c_{\ego}}
\newcommand{\adoinit}{c_{\ado}}
\newcommand{\egostepsize}{k}
\newcommand{\egostepsizetrain}{k_\trn}
\newcommand{\ntrain}{n_\trn}
\newcommand{\thetatrain}{\lambda_\trn}
\newcommand{\adorandom}{\ado[{\mathrm{rand}}]}
\newcommand{\adotrained}{\ado[\thetatrain]}

\newcommand{\initpositions}{C}

\newcommand{\muc}[2]{\multicolumn{#1}{c}{#2}}

\newcommand{\throttle}{\alpha}
\newcommand{\throttlemax}{\alpha_{\max}}
\newcommand{\throttlemin}{\alpha_{\min}}

\newcommand{\dist}{d}
\newcommand{\distlat}{d_\mathrm{lat}}
\newcommand{\distlong}{d_\mathrm{long}}
\newcommand{\dsafe}{d_\mathrm{safe}}
\newcommand{\pd}{u}
\newcommand{\vego}{v_\ego}
\newcommand{\vado}{v_\ado}
\newcommand{\disttwo}{d_2}
\newcommand{\dlka}{d_\mathrm{lka}}
\newcommand{\vlatado}{v_{\ado,\mathrm{lat}}}

\newcolumntype{M}{>{$\displaystyle}l<{$}} 
\newcommand\AddLabel[1]{\refstepcounter{equation}(\theequation)\label{#1}}
\newcolumntype{L}{>{\collectcell\AddLabel}r<{\endcollectcell}}

\newcommand{\dlego}{d_{\ell,\ego}}
\newcommand{\dlado}{d_{\ell,\ado}}
\newcommand{\elly}{\ell_Y} 
\newcommand{\ellr}{\ell_R} 

\newtheorem{definition}{Definition}
\newtheorem{theorem}{Theorem}
\newtheorem{lemma}{Lemma}

\allowdisplaybreaks

\title{Automatic Testing With \\Reusable Adversarial Agents}
\author{\IEEEauthorblockN{Xin Qin}
\IEEEauthorblockA{University of Southern California\\
Los Angeles, CA\\
xinqin@usc.edu}
\and
\IEEEauthorblockN{Nikos Ar\'echiga}
\IEEEauthorblockA{Toyota Research Institute\\
Los Altos, CA\\
nikos.arechiga@tri.global}
\and
\IEEEauthorblockN{Jyotirmoy Deshmukh}
\IEEEauthorblockA{University of Southern California\\
Los Angeles, CA\\
jdeshmuk@usc.edu}
\and
\IEEEauthorblockN{Andrew Best}
\IEEEauthorblockA{Toyota Research Institute\\
Los Altos, CA\\
andrew.best@tri.global}
}
\begin{document}
\maketitle

\begin{abstract}
    Autonomous systems such as self-driving cars and general-purpose
robots are safety-critical systems that operate in highly uncertain
and dynamic environments. We propose an interactive multi-agent
framework where the system-under-design is modeled as an ego agent and
its environment is modeled by a number of adversarial (ado) agents.
For example, a self-driving car is an ego agent whose
behavior is influenced by ado agents such as pedestrians, bicyclists,
traffic lights, road geometry etc. Given a logical specification of
the correct behavior of the ego agent, and a set of constraints that
encode reasonable adversarial behavior, our framework reduces the
adversarial testing problem to the problem of synthesizing controllers
for (constrained) ado agents that cause the ego agent to violate its
specifications. Specifically, we explore the use of tabular and deep
reinforcement learning approaches for synthesizing adversarial agents.
We show that ado agents trained in this fashion are better than traditional
falsification or testing techniques because they can generalize to ego
agents and environments that differ from the original ego agent.  We
demonstrate the efficacy of our technique on two real-world case
studies from the domain of self-driving cars. 

\end{abstract}

\section{Introduction}
\label{sec:intro}
Autonomous cyber-physical systems such as self-driving vehicles and
unmanned aerial vehicles operate in highly uncertain environments. For
such systems, it is imperative to develop a testing framework where
the system-under-design (SUD) is exposed to challenging scenarios
posed by its operating environment.  The view we adopt in this paper
is that an autonomous system and its environment can be modeled as an
interactive multi-agent system, where the SUD is the ego agent and the
environment can be viewed as a composition of several {\em
ado}\footnote{We use the term {\em ado} as an
abbreviation for adversarial agent.} agents.  For example, when
designing software for a self-driving car, the car itself is the ego
agent, while aspects such as other vehicles, pedestrians, traffic
lights, road markings, road curvature, and weather (among others) can
be considered as ado agents capable of autonomous behavior. The main
goal of an {\em adversarial testing} procedure is to find behaviors
presented by the ado agents that induce an undesired behavior by the
ego agent. An interesting challenge is that unrestricted ado agents
can lead to trivial violations of the ego agent's specification.  For
example, in the self-driving domain, consider an ego vehicle following
an ado car on a highway; a reasonable assumption is that the ado car
does not travel backward and that it follows traffic laws. Thus, the
goal of adversarial testing is actually to find violations of the ego
agent behavior subject to a set of constrained ado agents.

There is significant existing work that addresses adversarial testing.
The most prominent class of techniques focuses on requirement
falsification
\cite{staliro,breach,akazaki2018falsification,10.1007/978-3-030-25540-4_23,zhangHybridSystemFalsification2020}
; here, we assume that the SUD is a black-box model with time-varying
inputs and outputs, and its correctness specification is expressed as
a Signal Temporal Logic (STL) property of the outputs. Various
black-box optimization heuristics are used to search over the
(constrained) space of the model inputs to identify a violation of the
output property (See \cite{deshmukh2019formal} for a survey).
Search-based testing \cite{ali2009systematic} is another set of
related techniques that rely on heuristic search. Search-based testing
and falsification are fundamentally limited; these are {\em non}-{\em
adaptive} approaches, i.e., a counterexample is a fixed sequence of
actions undertaken by the environment (i.e. ado agents) which does not
generalize to new initial conditions for the agents or to changes in
the dynamics of the system. Thus, any change in the ego agent or the
environment necessitates re-running the falsification/search-based
testing tools, which is time-consuming in iterative design flows.
 
The recently developed adaptive stress testing (AST) approach
\cite{corso2019adaptive,koren2018adaptive} takes a different approach.
It uses a multi-agent view similar to the one we adopt in this paper
to find undesired behavior in ego agents. The goal in AST is not to
find a single (or a set of) counterexamples, but rather to learn {\em
an adversarial policy} (using reinforcement learning (RL))  that
causes the ego agent to fail. In this paper, we significantly extend
the AST idea in several ways. To better contextualize our
contributions, we use a motivating example.

\begin{figure}
    \centering
   \includegraphics[width=0.2\textwidth]{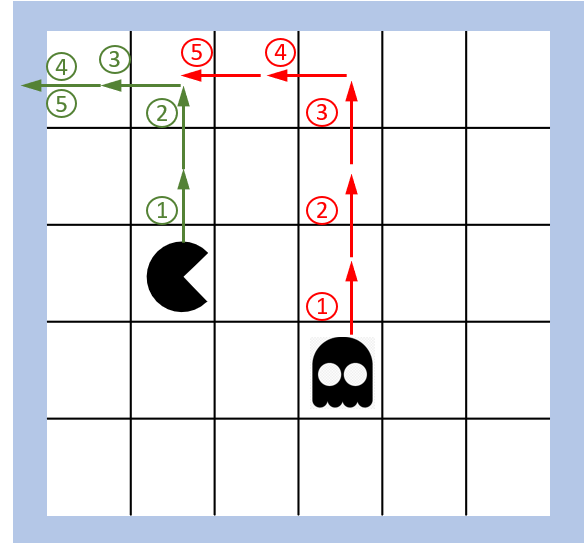} 
   \caption{Ego agent (pacman) evading adversarial agent 
             (ghost)\label{fig:gridworld}}
\end{figure}
\mypara{Motivating Example}
Consider two agents moving in a {\em grid}-{\em world} as shown in Fig.~\ref{fig:gridworld}, 
where at each time step, the ego agent moves one step in a direction away from the 
the adversary. If it picks a direction that has a grid boundary (wall), then it fails to 
move.  We say that the ``{\em ego is $r$-captured by the ado}'' if the Manhattan distance 
(denoted $\dman$) between the grid cells occupied by the ego (denoted $c_\ego$) and the 
ado (denoted $c_\ado$) is less than the positive integer $r$. In temporal logic, we can 
express the property that the ego not be captured for at least $k$ time steps 
as $\egospec$ $\equiv$ $\alw_{[0,k]} \left( \dman(c_\ego,c_\ado) > r\right)$. 
The goal is to find ado behavior that makes the ego violate this property. Consider a
motion model in which the ado agent can instantaneously jump to the location of the ego
agent -- clearly this is a trivial policy that ensures that the ego violates $\egospec$,
but this may not be reasonable behavior for the ado agent. Thus, there may be a need to
restrict the behaviors that the ado agent can have, and such constraints can also be
expressed using temporal logic. In this example, the constraint 
$\advrules$ $\equiv$ $\alw_{[0,k]} \left( \norm{v_\ado}_\infty \le 1 \right)$ 
restricts the maximum velocity of  the ado in either the X or Y direction to be less than 1.

Traditional testing tools 
\cite{staliro,breach} treat
falsification as a search problem over a user-defined parameter space. For example,
the user can introduce $N$ parameters for the $N$-length sequence of actions picked
by the ado agent. The space of actions $A$ can be discrete or continuous, and essentially
the falsification problem tries to find {\em one} sequence in $A^N$ such that the ego
violates $\egospec$. Fig~\ref{fig:gridworld} demonstrates such a sequence: the ado travels
upward 3 times and to the left twice, while the ego gets trapped in the top left
corner.  A sequence found by such a tool is however, {\em not generalizable}: suppose
the ego was already in the top left corner, and the ado was in the bottom right
corner, then blindly following the sequence (3 ups and 2 lefts) does not lead to
a violation of $\egospec$. Furthermore, if the ego were to allowed to move 2
cells in one time step (instead of 1), then such a fixed sequence will almost 
certainly not succeed in making the ego violate its spec. 

\mypara{Problem Definition} Given: (1) a formal specification for the ego
agent in a suitable temporal logic and a fixed implementation of its control
logic, (2) a set of (dynamic) constraints for each of the ado agents (also specified
in a suitable temporal logic), and (3) an executable model to simulate the
interactions between the agents, {\em the goal of this paper is to identify ado agent
behaviors that lead to a violation of the ego specification}.  Furthermore, 
we would like the testing strategy to be {\em robust to changes in the ego agent}, 
as outlined in the motivating example.  

\mypara{Solution Overview} While traditional adversarial testing tools focus on
searching over the action space of the ado agent(s), we propose a technique to
search over the {\em set of mappings from states to actions}, or the set of
policies for a given ado agent.  This can be done by parameterizing the control
logic of the ado agents, and using an appropriate controller synthesis procedure
to obtain ado {\em policies} that lead to ego agent violations. Specifically, we
explore  the use of {\em reinforcement learning} (RL). 
The work in \cite{koren2018adaptive,corso2019adaptive} also uses RL to identify
adversarial policies to make the ego agent display undesired behavior. There are
several key differences and improvements in our work: (1) In contrast to 
\cite{koren2018adaptive,corso2019adaptive}, where state-based rewards to be
used by the RL procedure are hand-created, we express both the
specification for the ego agent and the dynamic constraints for the ado agents
using temporal logic. This allows us to automatically infer the rewards 
to use in RL allowing us to avoid {\em reward engineering} -- a required but 
manual and error-prone step in RL \cite{rewardhacking}. (2) We show how our 
algorithms make the testing process robust -- our trained ado agents are able 
to find ego violations even after changes to the ego agent model {\em without 
having to re-train the ado agent for the perturbed ego agent}. Our procedure yields
reusable, robust and modular adversarial agents that can reduce the development 
time in an iterative design process.



\mypara{Summary of contributions}
\begin{enumerate}[wide, labelwidth=!, labelindent=0pt]

    \item We formulate a new adversarial testing algorithm for autonomous systems that
    views the SUD as an ego agent that interacts with a set of constrained ado agents. We
    assume correctness  specification for the ego agent, and the  behavioral constraints on 
    the ado agents are expressed in a formal logic such as Linear Temporal Logic (LTL) 
    \cite{ } or Signal Temporal Logic (STL) \cite{Maler2004}.
    We propose a framework where controller synthesis can be used to train an ado agent 
    policy to discover violations of the ego specification, and explore a specific controller
    synthesis procedure based on deep reinforcement learning. Specifically, we leverage 
    the {\em quantitative semantics} of STL to guide the deep RL based ado agent synthesis.
    
    \item We identify assumptions under which the learned ado policies are 
    {\em robust}. In particular we show that if the learned ado policy demonstrates a
    violation of the ego specification, then this policy will discover violations in an 
    ego agent that (1) starts from different initial configurations, and (2) has different 
    dynamics than the original ego agent. 
    
    \item
    We demonstrate the efficacy of our approach on two case-studies from the self-driving
    domain. We show that aspects such as other cars, traffic lights, pedestrians, etc. can 
    be modeled as ado agents. We consider (1) an adaptive cruise control example where the 
    leading car is modeled as an ado agent, and (2) a controller that ensures safety during 
    an ado car merging into an ego car's lane. 
\end{enumerate}

\noindent The rest of the paper is organized as follows. In Section~\ref{sec:prelim} we
provide the background and problem definition. We define rewards to be used by our RL-based
testing procedure in Sec.~\ref{sec:learning}. We show how the ado agents generalize
in Section~\ref{sec:generalizability}, and 
provide detailed evaluation of our technique in Sec.~\ref{sec:case_studies}.
Finally, we conclude with a discussion on
related work in Section~\ref{sec:related_conc}.

\section{Problem Statement and Background}
\label{sec:prelim}
We first introduce the formal
description of a multi-agent system as a collection of deterministic dynamical agents.

\begin{definition}[Deterministic Dynamical Agents] \label{def:dynagent}
An agent $\agent$ is a tuple
$(\astates,\actions,\transitions,\initstates, \policy)$, where
$\astates$ is a set of agent states, $\actions$ is the set of agent
actions, $\transitions$ is a set of transitions of the form
$(\astate,\action, \astate')$, where $\action \in \actions$,
$\initstates \subseteq \astates$ is a set of designated initial states
for the agent, and finally the policy\footnote{Our framework can
alternatively include stochastic dynamical agents, where
$\transitions$ is  defined as a distribution over
$(\astates\times\actions\times\astates)$, and the control policy
$\policy$ is a stochastic policy representing a distribution over
actions conditioned on the current state of the agent, i.e.
$\pi(\action\mid\astate)$. Also, states $\astates$ and actions
$\actions$ can be finite sets, or can be dense, continuous sets.}
$\policy$ is a function mapping a state in $\astates$ to an action in
$\actions$.
\end{definition}

A multi-agent system \(\system = \setof{\ego, \ado_1, \ldots,
\ado_k}\) is a set of agents, with a designated ego agent $\ego$, and
a non-empty set of adversarial agents $\ado_1,\ldots,\ado_k$. The
state-space of the multi-agent system can be constructed as a product
space of the individual agent state spaces, and the set of transitions
of the multi-agent system corresponds to the synchronous product of
the transitions of individual agents. The transitions of the
multi-agent system when projected to individual agents are consistent
with individual agent behaviors. A behavior trajectory for an agent is
thus a finite or infinite sequence
$(t_0,\astate_0),(t_1,\astate_i),\ldots$, where $\astate_i \in
\astates$ and $t_i \in \reals^{\ge 0}$. We use $\traj$ to denote a
trajectory variable, i.e. a function mapping $t_i$ to $\astate_i$,
i.e. $\traj(t_i) = \astate_i$.  In many frameworks used for simulating
multi-agent systems, it is common to consider timed trajectories with
a finite time horizon $t_N$, and a fixed, discrete time step, $\Delta
= t_{i+1}-t_i, \forall i$.

\mypara{Signal Temporal Logic}
Signal Temporal Logic (STL) \cite{Maler2004} is a formalism to describe properties of
real-valued, dense-time trajectories. STL formulas are evaluated over behavior 
trajectories. An atomic STL formula is a predicate of the form
$f(s) \sim c$, where $s$ is a trajectory variable, $f$ is a real-valued function from 
$\astates$ to $\reals$, $\sim$ is a comparison operator, i.e. $\sim\in \setof{<,\le,>,\ge}$ and 
$c \in \reals$. STL formulas are constructed recursively using the Boolean logical
connectives such as negations ($\neg$) and conjunctions ($\wedge$) and the temporal 
operator ($\until$). It is often convenient to define Boolean 
connectives such as disjunction ($\vee$), implication ($\Rightarrow$) 
using the usual equivalences for Boolean logic. It 
is also convenient to define temporal operators $\ev_I \varphi$ as shorthand for 
$\top \until_I \varphi$, and $\alw_I \varphi$ as shorthand for $\neg \ev_I \neg \varphi$.
Each temporal operator is indexed by the time interval\footnote{Traditional syntax of STL
permits intervals that are open on either or both sides; for signals over discrete-time steps,
this provision is not required. Furthermore, we exclude intervals that are not
bounded above as we intend to evaluate STL formulas on finite time-length traces.} 
$I$ of the  form $[a,b]$, where $a,b \in \reals^{\ge 0}$. 
%
%

STL has both Boolean semantics that recursively define the truth value of the satisfaction of 
an STL formula in terms of the satisfaction of its subformulas and quantitative semantics
that are used to map a trajectory and a formula to a real value known as the robust
satisfaction value or simply, the {\em robustness}. Intuitively, the robustness approximates 
the distance between a given signal $s$ and the set of signals satisfying the formula $\varphi$ 
\cite{Fainekos2009}. There are numerous definitions for quantitative semantics of STL, for example
\cite{malerMonitoringTemporalProperties2004}
\cite{jaksicQuantitativeMonitoringSTL2018}.
The 
actual definition to be used is irrelevant to this paper as long as it is efficiently
computable.  We will assume that the robustness has been clamped to be in an interval $[\rho_{min}, \rho_{max}]$, with $\rho_{min} = -\rho_{max} > 0$

\ignore{
The quantitative semantics of STL are defined using the 
function $\rob$ as follows:
\begin{eqnarray*}
\rob(\top,s,t) & = & \infty \\
\rob(f(s) \ge c, s, t) & = & f(s_t) - c \\
\rob(\neg \varphi, s, t) & = & -\rob(\varphi,s,t) \\
\rob(\varphi_1 \wedge \varphi_2, s,t) & = &
        \min(\rob(\varphi_1,s,t), \min(\varphi_2,s,t)) \\
\rob(\alw_\varphi, s,t) & = &
        \inf_{t' \in t\oplus I} \rob(\varphi,s,t') \\
\rob(\ev_\varphi, s,t) & = &
        \sup_{t' \in t\oplus I} \rob(\varphi,s,t') \\
\rob(\varphi_1 \until_I \varphi_2, s, t) & = &
        \sup_{t' \in t\oplus I} (\rob(\varphi_2,s,t') \wedge\\
        & &\inf_{t'' \in [t,t')} \rob(\varphi_1,s,t''))
\end{eqnarray*}
By convention, if the time \(t\) is unspecified, it is assumed to be zero. }

When evaluating an STL formula, each time step requires an application
of the functions \(f\) that appear in the formula. We denote a signal
$s$ satisfying a formula $\varphi$ at time $t$ by $s,t
\models\varphi$, and $s \models \varphi$ is by convention defined as
$s,0 \models \varphi$.  In
\cite{Fainekos2009,donzeRobustSatisfactionTemporal2010}, the authors
show that $\rob(\varphi,s) \ge 0 \implies s \models \varphi$, and
$\rob(\varphi,s) < 0 \implies s\not\models\varphi$.  Following the
notation in \cite{donzeRobustSatisfactionTemporal2010}, we call the
signal variables appearing in the formulas as {\em primary signals},
and the intermediate results that arise from function  applications as
\emph{secondary signals}.

\begin{definition}[Definition 2 of~\cite{donzeRobustSatisfactionTemporal2010}]
Let \(f_k(s)\) be any arithmetic function \(f_k : \astates \rightarrow \mathbb{R}\) that
appears in an STL formula \(\varphi\). We call the vector of variables in \(s\) the
\emph{primary} signals of \(\varphi\), and their images by \(f\) \emph{secondary} signals,
\(\{y_k\}\).
\end{definition}

Lemma~\ref{lem:robust_satisfaction_nearby_secondaries} tells us that if two signals have
``nearby values'' and also generate nearby secondary signals, then if one of them robustly
satisfies an STL formula, the other will also satisfy that formula. We first formalize the
notion of distance between signals. 
\begin{definition}[Distance between signals]
\label{eq:dist_signals}
Given two signals \(s\) and \(s'\) with identical value domains $S$ and identical time
domains \(\mathbb{T}\), and a metric \(d_S\) on \(S\), the distance between \(s\) and \(s'\), denoted as \(\norm{s-s'}\) is defined as:
\(
\sup_{t\in\mathbb{T}} d_S(s(t), s'(t))
\).
\end{definition}

\begin{lemma}[Theorem 1 in~\cite{donzeRobustSatisfactionTemporal2010}]
\label{lem:robust_satisfaction_nearby_secondaries} 
If \(\rob(\varphi, s, t) = \delta\), then for every signal \(s'\) s.t.
every secondary signal satisfies \(\norm{y_k - y_k'} < \delta\), then 
\((s \models \varphi) \implies (s' \models \varphi)\).
\end{lemma}


\mypara{Problem Definition: Testing with Dynamically Constrained Adversaries}
Given a behavior of the multi-agent system, let the
projection of the behavior onto agent $\agent$ be denoted by the signal variable
$s_\agent$. Formally, the problem we wish to solve can be stated as follows:
\begin{enumerate}[leftmargin=1.3em,topsep=2pt,itemsep=2pt]
    \item Given a spec $\egospec$ on the ego agent,
    \item Given a set of constraints $\advrules_i$ on adversarial agent $\agent_i$,
    \item Find a multi-agent system policy that can generate behaviors such that:
\(\forall i:s_{\ado_i} \models \advrules_i \wedge s_\ego \not\models \egospec\).
\end{enumerate}
In other words, we aspire to generate a compact representation for a
possibly infinite number of counterexamples to the correct operation
of the ego agent.

\subsection{Adversarial Testing through Policy Synthesis}
In contrast to falsification approaches,
we assume a deterministic 
(or stochastic) dynamical agent model for the ado agents (as defined in Def.~
\ref{def:dynagent}), i.e. the $i^{th}$ ado agent is specified as a tuple of 
the form \((\astates_i,\actions_i,\transitions_i,\initstates_i,\policy_i)\). 
We  assume that initially all agents have a randomly chosen policy \(\policy_i\). 
For ado agent \(\ado_i\), let \(\policies_i = \actions_i^{\astates_i}\)  
denotes the set of all possible policies. Let $\policies_i(\advrules{i})$ be 
the set of policies such that for any $\policy \in \policies_i(\advrules{i})$,
using $\policy$ guarantees that the sequence of states of $\ado_i$ satisfies
$\advrules_i$. Similarly, let $\policies_i(\neg\egospec)$ be the set of policies
that guarantees that the sequence of states for the ego agent $\ego$ does
not satisfy $\egospec$. The problem we wish to solve is: for each $i$, find 
a policy in $\policies_i(\neg\egospec) \cap \policies_i(\advrules{i})$. 

One approach to solve this problem is to use a reactive synthesis approach, 
when specifications are provided in a logic such as LTL or ATL \cite{bloem2018graph,dimitrova2014deductive,liu2013synthesis}.
There is limited work on reactive synthesis
with STL objectives \cite{raman2015reactive,ghosh2016diagnosis}, mainly requiring
encoding STL cosntraints as Mixed-Integer
Linear Programs; this may suffer from scalability in multi-agent
settings. We defer detailed comparison with reactive synthesis approaches to
future work. In this paper, we propose using the framework of deep reinforcement 
learning (RL) for controller synthesis with a procedure for automatically 
inferring rewards from specifications and constraints. 

\subsection{Policy synthesis through Reinforcement Learning}
Reinforcement learning (RL) \cite{sutton_reinforcement_2018} and related deep
reinforcement learning (DRL) \cite{mnih_playing_2013} are procedures to train
agent policies in deterministic or stochastic environments. In our setting,
given a multiagent system \(\system = \setof{\ego, \ado_1, \dots, \ado_k}\),
we wish to synthesize a policy \(\policy_k\) for each ado \(\ado_k\). We can 
model multiple ado agents as a single agent whose state is an element
of the Cartesian product of the state spaces of all agents, i.e.,
\(\astates = \astates_\ego \times \astates_{\ado_1} \times \dots \times \astates_{\ado_k}\), 
and the action of this single agent is a a tuple of actions of all 
{\em ado} agents, i.e. \(\actions = \actions_{\ado_1} \times \ldots \times \actions_{\ado_k}\).

In each step, we assume that the agent is in 
state $\astate \in \astates$  and interacts with the environment by taking
action $\action \in \actions$. In response, the environment (i.e. the
transition relation of the ado and the ego agents) picks a next state $\astate'$ 
s.t. $(\astate,\action,\astate') \in \transitions$,  and a reward 
$\reward(\astate,\action)$. The reward provides reinforcement for the 
constrained adversarial behavior, and will be elaborated in Section~\ref{sec:learning}. 
The goal of the RL agent is to learn a deterministic policy $\policy(\astate)$, 
such that the long term payoff of the agent from the initial state (i.e. 
the discounted sum of all rewards from that state) is maximized. As is common
in RL, we define the notion of a value function $\valuefn$ in
Eq.~\eqref{eq:value}; this is the expected reward over all possible actions
that may be taken by the agent. In a deterministic environment, the 
expectation disappears.
\begin{equation}
\label{eq:value}
\valuefn_\policy(\astate_t) = \expectedvalue_{\policy}\left[
    \begin{array}{ll}
    \displaystyle\sum_{k=0}^{\infty}  & \gamma^k
        \reward((\astate_{t+k}),\action_{t+k})\ \Bigg\lvert
     \action_{t+k} = \policy(\astate_{t+k})
    \end{array}\right]
\end{equation}
RL algorithms use different strategies to find an {\em optimal} policy
$\policy*$, that for all $\astate$ is defined as $\policy^*(\astate) =
\argmax_{\pi} \valuefn_\policy(\astate)$.  We assume that the state of
the agent  at time $t_0$, i.e. $\astate_0$ is in $\initstates$. 

We now briefly review a classic model-free RL algorithm called q-learning and
summarize two deep RL algorithms: Deep Q-learning and Proximal Policy Optimization (PPO). 
In q-learning, we learn a state-action value function 
\(q(\astate, \action)\), which represents the believed value of taking action 
\(\action\) when in state \(\astate\). Note that 
\(\valuefn(\astate) = \max_\action q(\astate, \action)\). 
In q-learning, the agent maintains a table whose rows correspond to the states of the
system and columns correspond to the actions. The
entry \(q(\astate, \action)\), 
encodes an approximation to the state-action value function computed by the algorithm. 
The table is initialized randomly. At each time step \(t\), the agent 
uses the
table to select an action \(\action_t\) based on an
\(\varepsilon\)-greedy policy, i.e. it chooses a random action with
probability $\varepsilon$, and with probability $1-\varepsilon$, 
chooses $\text{argmax}_{\action \in \actions} q(\astate,\action)$. 
Next, at time step \(t+1\), the agent observes the reward received
\(R_{t+1}\) as well as the new state \(\astate_{t+1}\), and it uses this
information to update its beliefs about its previous behavior.
During the training process, 
we sample the initial state of each episode with a probability distribution \(\mu(\astate)\) 
that is nonzero at all states. Given enough time, all states will 
eventually be selected as the initial state. We also fix the policies of 
the ado agents to be \(\varepsilon\)-soft. This means that for each state \(\astate\) 
and every  action \(\action\), \(\pi(a|x) \ge \varepsilon\), where \(\varepsilon > 0\) is a 
parameter. Together, random sampling of initial states and \(\varepsilon\)-soft 
policies ensure that the agent performs sufficient exploration and avoids converging 
prematurely to local optima. 



%

\mypara{Deep RL} Deep RL is a family of algorithms that make use of Deep Neural Networks (DNNs) to represent either the value or the policy of 
an agent. Deep Q-learning 
\cite{mnih_playing_2013}, is an extension of q-learning where the table \(q(s,a)\) is approximated by a
DNN, \(q(s,a,w)\), where \(w\) are the network parameters.
Deep Q-learning observes states and selects actions similarly to Q-learning, but it additionally uses 
\emph{experience replay}, in which the agent stores previously observed tuples of states, 
actions, next states, and rewards. At each time step, the agent updates its q-function 
with the currently observed experience as well as with a batch 
of experiences sampled randomly from the experience replay buffer\footnote{Tabular Q-learning is guaranteed to converge to the optimal value function 
\cite{sutton_reinforcement_2018}. On the other hand, DQN may not converge, but it will eventually find a counterexample trace if it exists. In practice, 
DQN performs well and finds effective value functions, even if its convergence cannot be theoretically guaranteed.}.
The agent
then updates its approximation network by gradient descent on the quadratic loss 
function
\(
\mathcal{L} = (y_t - q(\astate_t, \action_t, w))^2
\),
where
\(
y_t = R_{t+1} + \gamma \max_{\action'} q(\astate_{t+1}, \action, w).
\)
In the case that \(\astate_t\) is a terminal state, it is common to assume
that all transitions are such that \(\astate_{t+1} = \astate_t\) and \(R_t = 0\).
Although these learning algorithms learn the state-action value function \(q(\astate,\action)\),
in the theoretical exposition that follows, we will use the \emph{state value function} 
\(\valuefn(\astate)\) for simplicity. The optimal state value function can be 
obtained from the optimal state-action value function by \(\optimalvalfcn{\astate} =
\max_\action q^{\star}(\astate,\action)\).
PPO \cite{schulman2017proximal} is a state-of-the-art {\em policy gradient} algorithm 
that performs gradient-based updates on the policy space directly while ensuring that the new
policy is not too far from the old policy.


\section{Learning Constrained Adversarial Agents}
\label{sec:learning}
In this section, we describe how we construct a reward function that enables
training constrained adversarial agents. Note that the reward function needs
to encode two aspects: (1) satisfying ado constraints, (2) violating
ego specification. We assume that ado constraints are hierarchically ordered
with priorities, inspired by the Responsibility-Sensitive Safety rules for
traffic scenarios in \cite{censi_liability_2019}.

\begin{definition}[Constrained adversarial reward]\label{defn:stlreward}
Suppose from initial state
\(\astate_0\) the agent has produced a behavior trace \(\traj(\astate_0)\) of duration \(T\). We distinguish two cases.
\begin{itemize}[align=left,leftmargin=0pt, labelindent=0pt, listparindent=\parindent, itemindent=!]
    \item Case 1: All ado constraints are strictly satisfied, i.e. \(\rob(c, \traj(\astate_0))> 0\) for each constraint \(c \in \advrules\). In this case, the reward will be the robustness of the adversarial specification.
    \begin{align}
        \reward_t =   
         \left\{  \begin{array}{ll}  0 
          & \text{if $t < T$} \\
        \rob(\neg\egospec,\trajsig{\astate_0}) & \text{ if $t = T$}
    \end{array}
    \right.
    \end{align}
    \item Case 2: Not all ado constraints are strictly satisfied. Let \(c\) be the highest priority rule that is not strictly satisfied, i.e. the highest  priority rule such that \(\rob(c,\traj(\astate_0)) \le 0\).  Then, every constraint with priority higher than \(c\) will contribute zero, whereas every constraint with priority less then or equal to \(c\) will contribute \(\rho_{min}\). Let \(M\) be the number of constraints with priority less than or equal to \(c\).
        \begin{align}
        \reward_t = \left\{  \begin{array}{ll}
        0 & \text{if $t < T$} \\
        - M \rho_{min} & \text{if $t = T$}
        \end{array}
    \right.
    \end{align}
\end{itemize}
\end{definition}

The following lemma shows that it is not possible for an ado agent to
attain a high reward for satisfying
lower priority constraints at the expense of higher priority
constraints. The proof is straightforward (shown in the appendix).
\begin{lemma}[Soundness of the reward function]\label{lem:reward_soundness}
Consider two traces, \(\traj_1\) and \(\traj_2\). 
Suppose that the highest priority constraint violated by 
\(\traj_1\)
is
\(c_1\) and the highest priority constraint violated by
\(\traj_2\)
is \(c_2\). Suppose
\(c_1\) has lower priority than \(c_2\). Then,
the reward for trajectory 1 will be higher than the reward for trajectory 2, i.e.
\(\reward(\traj_1) > \reward(\traj_2)\).
\end{lemma}

\section{Generalizability of Adversarial Agents}
\label{sec:generalizability}

In this section we show how our RL-based testing approach makes the testing procedure
itself robust by learning a closed-loop policy for testing. 
We first show generalizability across initial conditions in two steps: (1) In Theorem~\ref{thm:generalization_initial_conditions}, we assume 
that the RL algorithm has converged to the optimal value function, 
and that it has used this value
function to find a counterexample trace \(\traj(\astate_0)\). We
consider a new state \(\astate_0'\), and want to bound the degradation of
the robustness function of the specification \(\rob(\neg\egospec,\traj(\astate_0'))\). (2) In Theorem~\ref{thm:relax}, we {\em relax} this strong assumption and identify conditions under which
generalization can be guaranteed even with approximate convergence. In Theorem~\ref{thm:bisim}, we 
show generalizability when the ego agent dynamics change. The main idea is that if the new
dynamics have an $\epsilon$-approximate bisimulation relation to the original dynamics,
then we can guarantee generalizability.


\begin{theorem}\label{thm:generalization_initial_conditions}
Suppose that the adversarial agent has converged to the optimal value function \(\optimalvalfcn{\astate}\), and that it has found a trace \(\traj(\astate_0)\) that falsifies the target specification \(\egospec\) with robustness 
\(\rob(\neg\egospec, \traj(\astate_0)) = \tau > 0\) while satisfying all of the ado constraints. Given a new state \(\astate_0'\) such that \(\absval{\optimalvalfcn{\astate_0} - \optimalvalfcn{\astate_0'}} < \delta\) with \(\delta < \discountfactor^T |\rob_{min}|\), the adversary will be able to find a new trajectory \(\traj(\astate_0')\) that satisfies all of the constraints. Furthermore, the robustness of the specification \(\neg\egospec\) over the new trace will be at least \(\tau - \delta/\discountfactor^T\).
\end{theorem}

\begin{proof}
We can expand the optimal value function at state \(\astate_0\) as \(\optimalvalfcn{\astate_0} = \sum_{t=0}^T \discountfactor^t \reward_t\), where \(\reward_t\) is the reward function defined in Definition \ref{defn:stlreward}. Then, the optimal value function at \(\astate_0\) is \(\optimalvalfcn{\astate_0}  = \discountfactor^T \tau\).
Suppose for a contradiction that the new trajectory \(\traj(\astate_0')\) violates some number $M$ of constraints. Then, the following equations show that the two states must actually differ by a large amount, much larger than $\delta$, leading to a contradiction.
Formally, from Definition \ref{defn:stlreward} we have
\begin{align}
    \optimalvalfcn{\astate_0'} &= -\discountfactor^T M \rob_{min},\\
    \absval{\optimalvalfcn{\astate_0'} - \optimalvalfcn{\astate_0}} &\ge \discountfactor^T \tau + M \discountfactor^T \rob_{min} > \delta,
\end{align}
which contradicts the assumption of the theorem.  As the constraints will be 
satisfied by \(\traj(\astate_0')\), their contribution to the value function at
\(\astate_0'\) will be zero. Then, for the $2^{nd}$ part of the theorem we can 
expand the optimal value function as:
\begin{align}
\absval{\optimalvalfcn{\astate_0}\!-\!\optimalvalfcn{\astate_0'}}\!=\!
 \absval{\discountfactor^T \rob(\neg\egospec,\traj(\astate_0))\!-\!
 \discountfactor^T \rob(\neg\egospec,\traj(\astate_0'))} 
\end{align}
By assumption the above terms are $\leq \delta$, which gives us that
\begin{align}
&\absval{\rob(\neg\egospec,\traj(\astate_0)) - \rob(\neg\egospec,\traj(\astate_0'))) } \le \frac{\delta}{\discountfactor^T}
\end{align}
and the theorem follows.
\end{proof}
Note that if $\delta$ is chosen as a small enough perturbation such that $\tau - \delta/\discountfactor^T > 0$, then the new trace is also a trace in which the adversary causes the ego to falsify its specification.


The tabular Q-learning algorithm is guaranteed to converge asymptotically to the optimal value
function, meaning that for any \(\epsilon\), there exists a \(k\) such that at the \(k\)-th 
iteration, the estimate \(V_k\) differs from the optimal value function by at most \(\alpha\), 
i.e. \(\forall \astate \in \astates\), \(\absval{\optimalvalfcn{\astate} - V_k(\astate)} < \alpha\). 
Some RL algorithms have even stronger guarantees. For example, Theorem 2.3 of
\cite{fernsBisimulationMetricsContinuous2011}, states that running the value iteration algorithm
until iterates of the value function differ by at most 
\(\frac{\alpha (1-\discountfactor)}{2 \discountfactor}\) produces a value function that 
converges within \(\alpha\) of the value function \(\absval{\suboptimalvalfcn{\astate_0} -
\optimalvalfcn{\astate_0}} \le \alpha\). While useful for theoretical results, value iteration 
does not scale to problems with large state spaces. The following theorem states that if the RL 
algorithm has found a value function that is near optimal, the agent will be able to generalize
counterexamples across different initial states.

\begin{theorem}
\label{thm:relax}
Suppose that we have truncated an RL algorithm at iteration \(k\). Suppose that, from the guarantees of this particular RL algorithm, we are within \(\alpha\) of the optimal value function, i.e. for every \(\astate\), \(\absval{\suboptimalvalfcn{\astate_0} - \optimalvalfcn{\astate_0}} \le \alpha\). Further suppose that the adversarial agent has found a falsifying trace from state \(\astate_0\) with robustness \(\tau\), i.e. \(\rob(\neg\egospec,\traj(\astate_0)) = \tau\).
Now consider a new state \(\astate_0'\) such that the degradation of our approximate value function is at most \(\beta\), i.e.
\(    \absval{\suboptimalvalfcn{\astate_0} - \suboptimalvalfcn{\astate_0'}} \le \beta
\). Then, the adversarial agent will be able to produce a new trace with robustness at least
\begin{align}
    \rob(\neg\egospec,\traj(\astate_0'),T) \ge \tau - \frac{2 \alpha + \beta}{\discountfactor^T}
\end{align}
\end{theorem}
\begin{proof}
Note that by the triangle inequality
\begin{align}
\alpha + \beta >& 
\absval{\suboptimalvalfcn{\astate_0} - \optimalvalfcn{\astate_0}} + 
\absval{\optimalvalfcn{\astate_0'} - \suboptimalvalfcn{\astate_0}}\\ >& 
\absval{\suboptimalvalfcn{\astate_0'} - \optimalvalfcn{\astate_0}} \end{align}
Further,
\begin{equation}
    2 \alpha + \beta >
        \absval{\suboptimalvalfcn{\astate_0'} - 
        \optimalvalfcn{\astate_0}} >
        \absval{\optimalvalfcn{\astate_0'} - 
        \optimalvalfcn{\astate_0}}
\end{equation}
The result now follows from Theorem \ref{thm:generalization_initial_conditions} by substituting 
\(\delta = 2\alpha + \beta\).
\end{proof}

\noindent Finally, we will show that the agent may cope with limited changes to the multiagent 
system. This is useful in a testing and development situation, because we would like to be
able to reuse a pre-trained ado to stress-test small modifications of the ego without 
expensive retraining. To do this, we will define an \(\epsilon\)-bisimulation relation 
that will allow us to formally characterize the notion of similarity between different 
multiagent systems.
\begin{definition}[\(\epsilon\)-approximate bisimulation, \cite{girardApproximateBisimulationBridge2011}]
Let \(\epsilon > 0\), and \(\system_1\), \(\system_2\) be systems with state spaces \(\astates_1\), \(\astates_2\) and transition relations \(\transitions_1\), \(\transitions_2\), respectively. A relation \(\mathcal{R}_{\epsilon} \subseteq \astates_1 \times \astates_2\) is called an \(\epsilon\)-approximate bisimulation relation between \(\transitions_1\) and \(\transitions_2\) if for all \(\astate_1, \astate_2 \in \mathcal{R}_{\epsilon}\),
\begin{enumerate}
    \item $d(x_1, x_2) \le \epsilon$ where $d$ is a distance metric
    \item $\forall \action \in \actions$, $\forall \astate_1' \in \transitions_1(\astate_1, \action)$, $\exists \astate_2' \in \transitions_2(\astate_2, \action)$ such that $(\astate_1', \astate_2') \in \mathcal{R}_{\epsilon}$
    \item $\forall \action \in \actions$, $\forall \astate_2' \in \transitions_2(\astate_2, \action)$, $\exists \astate_1' \in \transitions_1(\astate_1, \action)$ such that $(\astate_1', \astate_2') \in \mathcal{R}_{\epsilon}$
\end{enumerate}
\end{definition}

\begin{theorem}
\label{thm:bisim}
Suppose the adversarial agent has trained to convergence as part of a multiagent 
system \(\system_1\), and it has found a trace that satisfies the adversarial specification
with robustness \(\tau\). Consider a new multi-agent system \(\system_2\) and suppose there 
exists an \(\epsilon\)-approximate bisimulation relation between the two systems, including 
the secondary signals of the formula \(\neg\egospec\). Further suppose that $\epsilon < \tau$ Then, the trajectory of the new system 
will also violate the ego specification while respecting the ado constraints.
\end{theorem}

\begin{proof}
If there is an \(\epsilon\)-approximate simulation between the primary and secondary signals 
of the traces of the two systems, then for a trace \(\traj_1(\astate_0)\) of system \(\system_1\)
starting from initial state \(\astate_0\), and a trace \(\traj_2(\astate)\) of system \(\system_2\)
also starting from initial state \(\astate_0\), the \(\epsilon\)-approximate bisimulation
relation ensures that both the primary and secondary signals differ by at most epsilon, i.e.
\(   (\absval{\traj_1(\astate_0) - \traj_2(\astate_0)} \le \epsilon) \wedge (\absval{y_1 - y_2} \le \epsilon).\)
From Lemma \ref{lem:robust_satisfaction_nearby_secondaries} it follows that \(\traj_2\) also causes the ego to falsify its specification while satisfying the
ado specifications.
\end{proof}


\section{Case Studies}
\label{sec:case_studies}
In this section, we use the motivating example introduced in
Section~\ref{sec:intro} to first empirically demonstrate the
robustness of our adversarial testing procedure. Then, we demonstrate
scalability of adversarial testing by applying it to two case studies
from the autonomous driving domain. 

\subsection{Benchmarking Generalizability}

In the grid world example from Section~\ref{sec:intro}, the
environment consists of an $n \times n$ grid containing the
ego agent and the ado agent. The objective of the ego agent is to
escape ado agents, assuming that the game begins with 
the ego and the ado agent at $(\egoinit,\adoinit)$.
The ego agent can move $\egostepsize$ cells in
any time step. The ego specification and ado constraints are as specified 
in Section~\ref{sec:intro}.


The ego policy is hand-crafted: it observes the ado position and selects the
direction (up, down, left, or right) that maximizes its distance from the ado.
If the target cell lies outside the map, it chooses a fixed direction to move away.
It is important that it is not trivial for the ado agent to capture
the ego agent; thus, for every experiment, we provide
a baseline comparison with an ado agent that has a randomly chosen
policy. A random policy has some likelihood of succeeding from a given
initial configuration of the ego and ado agents. Thus, the ratio of
the number of initial conditions from which the random ado agent
succeeds to the total number of initial conditions being tested
quantifies the {\em degree of difficulty} for the experiment. We
denote the random ado agent as $\adorandom$. 


In all the experiments in this section, we train the ado agent using
our RL-based procedure on a training arena characterized by the vector
$\thetatrain = (\ntrain, \initpositions, \egostepsizetrain)$, i.e.  a
fixed grid size ($\ntrain\times\ntrain$), a set of initial positions
$(\egoinit,\adoinit) \in \initpositions$, and (3) a fixed step size
for the ego agent movement ($\egostepsizetrain$).  We use Proximal
Policy Optimization (PPO)-based deep-RL algorithm to train the ado
agent\cite{schulman2017proximal}. 
We denote this trained agent as $\adotrained$
for brevity.  We frame the empirical validation of our robust
adversarial testing in terms of the following research questions:

\begin{enumerate}[label={\textbf{RQ\arabic*}.},align=left,leftmargin=0pt, 
labelindent=0pt, listparindent=\parindent, itemindent=!]
\item
How does the performance of $\adotrained$ compare against $10$
uniformly sampled $\adorandom$ agents on the same set of initial
positions used to train $\adotrained$ when all other arena parameters
remain the same? [To demonstrate degree of difficulty.]
\item
How does the performance of $\adotrained$ compare against
$\adorandom$ in an arena of varying map sizes when all other parameters remain the same?
\item
How does the performance of $\adotrained$ compare against
$\adorandom$ in an arena with varying ego agent step-sizes where all other
parameters remain the same?
\item
Does the ado agent generalize across initial conditions, i.e. if
we pick a small subset of the initial conditions to train an ado
policy, does the policy discover counterexamples on initial states that
were not part of the training set?
\item
Does the ado agent generalize to arenas with $\epsilon$-bisimilar dynamics?
\end{enumerate}

\begin{table}[h]
\begin{tabular*}{.49\textwidth}{l@{\extracolsep{\fill}}cll}
\toprule
Experiment           & Parameter       & \muc{2}{Success Rate (\%)} \\
\cmidrule(l){3-4} 
                     &                 & $\adotrained$  & $\adorandom$  \\
\midrule
                     & Num. $\adorandom$ &                    &                 \\
\cmidrule(l){2-4}
\multirow{2}{*}{RQ1} &  10             &   67.92 (163/240)  & 9.83 (237/2400)   \\  
                     &  20             &   67.92 (163/240)  & 10.23 (491/4800)  \\
\midrule
\multirow{4}{1em}{RQ2} & Map Size      &              &             \\
\cmidrule(l){2-4}
                     & $2\times 2$     & 66.67    (8/12)       &  33.33  (4/12)           \\
                     & $\mathbf{4\times 4}$     & 67.92  (163/240)  &  2.91(7/240)             \\
                     & $5\times 5$     & 70.33  (422/600)       &    1.0 (6/600) \\
                     & $10\times 10$   & 9.26  (917/9900)      &  0.33  (33/9900)   \\
\midrule
\multirow{4}{*}{RQ3} & Ego Step-size   &              &             \\
\cmidrule(l){2-4}
                     & $4$            &   72.50 (174/240)    &  2.92  (7/240)          \\
                     & $3$            &   72.50 (174/240)   &      2.92  (7/240)       \\
                     & $\mathbf{2}$   &67.92  (163/240)   &      2.92  (7/240)        \\
                     & $1$            &    67.92  (163/240) &   7.08  (17/240)           \\
\midrule
\multirow{4}{*}{RQ4} & $\delta$   &  \muc{2}{Avg. Success Rate}     \\
\cmidrule(l){2-4}
                     &  1        &     \muc{2}{0.9735}  \\
                     &  0.5      &     \muc{2}{0.9742}  \\
                     &  0.1      &     \muc{2}{0.9883}  \\
                     &  0.01     &     \muc{2}{1.0}     \\
\midrule
\multirow{2}{*}{RQ5} & $\epsilon$   &  \muc{2}{Avg. Success Rate}     \\
\cmidrule(l){2-4}
                     & 3            &   \muc{2}{0.89} \\
                     & 5            &   \muc{2}{0.78} \\
\bottomrule
\end{tabular*}
\caption{Empirical demonstration of the robustness of adversarial
testing.\label{tab:transfer}}
\end{table}

For the first three RQs, we use $\thetatrain = (4,C,2)$, where $C$ is the set of all
possible initial conditions for the agents.
Table~\ref{tab:transfer} shows the results for RQ1. Our trained
agent easily surpasses the average performance of both $10$ and $20$
random ado agents across all initial locations in the training set
$C$. The average number of violations found by a random ado agent
is around 10\%, while our trained ado agent captures the ego within 
the given time limit from 68\% of the initial conditions. Thus, finding an
ado policy that works for a majority of the initial cells is
sufficiently difficult.  From the results for RQ2, we observe that
$\adotrained$ is successful at causing the ego agent to violate its
specification for varying map sizes, even as large as $10\times 10$,
though it was trained on a $4 \times 4$ map. In contrast a random
ado agent is rarely successful. Finally, from the results for RQ3, we
observe that the trained ado agent succeeds even against an ego that 
uses different step sizes than those on which the ado was trained.

For RQ4, we used $\thetatrain = (10,C_{100},1)$, where $C_{100}$ is a set
of 100 randomly sampled initial positions (note that the total number
of initial configurations is 4950). We found $100$ counterexample states
during training, of which we chose $5$ states at random. For each of these
states $\astate$, we obtained the value of  
the state as maintained by the PPO algorithm, and found all states $\astate'$ 
s.t. $\absval{\valuefn(\astate)-\valuefn(\astate')} < \delta$. We then
computed the fraction of these states that also led to counterexamples.
For four of the identified counterexample states, $\delta$ satisfied
the conditions outlined in Theorem~\ref{thm:generalization_initial_conditions}, and
the results are shown in Table~\ref{tab:transfer}. As expected, smaller the
value of $\delta$, higher is the number of failing states with nearby
values. 

For RQ5, we used $\thetatrain = (10,C_{200},1)$, where $C_{200}$
was a randomly chosen set of 200 initial states. Then we defined a
refinement of the map, basically an 
$\epsilon \times \epsilon$ grid was imposed on each grid cell of the
original map. We considered an ado policy that basically used the
same action as that of the original coarser grid cell, while the
ego agent used a refined policy. We can establish that the transition
system thus engendered is actually $\epsilon$-bisimilar to the original
transition system. For different values of $\epsilon$, we picked $3$ 
sets of $300$ random initial states and tested if they led to counterexamples.
The average success
rates are shown in Table~\ref{tab:transfer}. We see that an abstract
ado policy can also violate the ego spec surprisingly often.

\subsection{Autonomous Driving Case studies}
We now apply our adversarial testing framework to two case studies
from the autonomous driving domain\footnote{We provide one more case 
study in the appendix.}. The case studies were implemented
in the Carla driving simulator \cite{Dosovitskiy17} as a means to
stress-test a controller driving a car in two different scenarios,
(1) freeway driving on a straight lane, 
(2) freeway driving with a car
merging into the ego lane. The ado agent was developed in
\texttt{python}, and the neural networks used in the DQN examples were
developed in \texttt{pytorch} \cite{pytorch}.
\begin{table}[t]
\begingroup\setlength{\fboxsep}{0pt}
\begin{tabular}{llML}
\toprule
Case Study & Description & \muc{2}{STL Formula}  \\
\midrule
\multirow{2}{*}{ACC} & Ego: Avoid collision & 
\alw_{[0,T]} (\dist \ge \dsafe) & eq:accegospec \\
                     & Ado: Velocity bounds & 
\alw(v_{\min} \le \vado \le v_{\lim}) & eq:accadorule\\
\midrule
Lane  & Ego: Avoid collision & 
\alw_{[0,T]} (\disttwo \ge \dsafe) & eq:lcegospec \\
Change & Ado: Init. pos.& 
\distlong > \dsafe & eq:lcadorule \\
\bottomrule
\end{tabular}
\endgroup
\caption{Ego Specifications and Ado Rules for case studies\label{tab:specsandrules}}
\end{table}

\begin{figure}[t]
\centering
\subfloat[]{%
\includegraphics[width=0.245\textwidth]{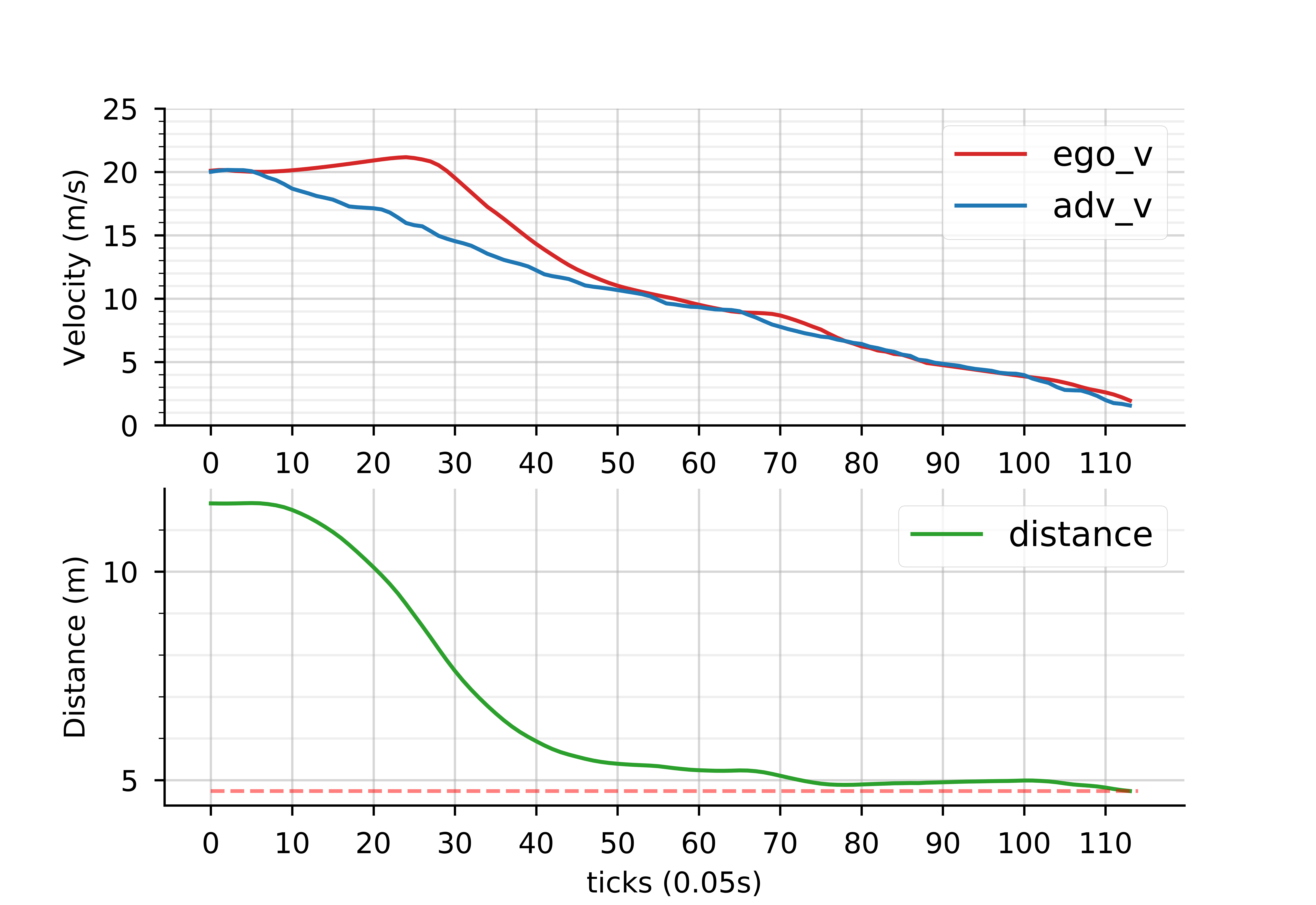}\label{traj_tb2}}%
\subfloat[]{%
\includegraphics[width=0.245\textwidth]{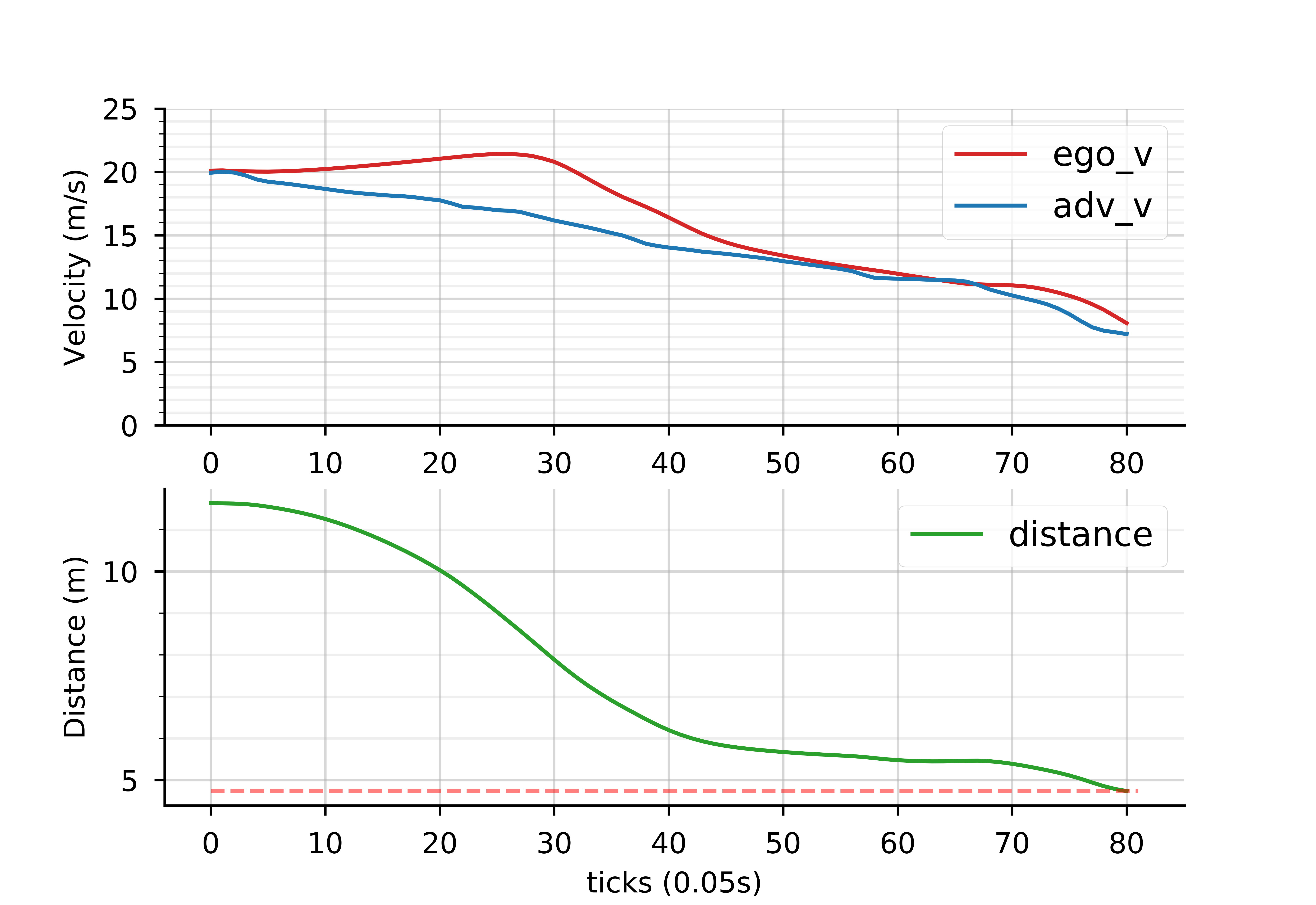}\label{traj_tb3}}
\caption{Traces showing two different ado policies that are able to
induce a collision in the ego agent.\label{fig:training_case_tb}}
\end{figure}

\mypara{Adaptive Cruise Control} In this experiment, two vehicles are
driving in a single lane on a highway.  The lead vehicle is the ado.
The follower vehicle avoids colliding into the leader using
an adaptive cruise controller (ACC). The purpose of adversarial
testing is to find robust ado policies that cause the ACC system to
collide with the ado.  The ACC controller modulates
the throttle ($\throttle$) by observing the distance ($\dist$) to the
lead vehicle and attempting to maintain a minimum safe following
distance $\dsafe$. The ACC controller is a
Proportional-Derivative (PD) controller with saturation. The PD term
$\pd$ is equal to 
 $K_p (\dist - \dsafe) + K_d (\vado - \vego)$, and the controller 
 action $\throttle$ is defined as $\throttlemax$ if $\pd > \throttlemax$,
$\throttlemin$ if $\pd < \throttlemin$, and $\pd$ otherwise.
The ego specification is given in Eq.~\eqref{eq:accegospec}. Here, $T$
is the maximum duration of a simulation episode and $\dsafe$ is the
minimum safe following distance. The ado should cause
the ego to violate its spec in less than $T$ seconds. The 
ado constraint specifies that it should not exceed the speed
limit $v_{\lim}$ and that it should maintain a minimum speed
$v_{\min}$. For our experiment, we choose $v_{\min} = 0.1$, and
$\dsafe = 4.7$m.  The distance $\dist$ is computed between the two
front bumpers. This represents a car length of \(4.54 m\), plus a
small safety margin.  The state of the ado agent is the tuple \(\dist,
\vego, \vado\).  At each time step, the ado agent chooses an
acceleration from a discretized space which contains \(3\)
possible actions. In this experiment, we explore two different RL
algorithms: q-learning and a DQN algorithm with replay buffers
\cite{Mnih2013PlayingAW}.  The average runtime using
the DQN (1.93 hours) is less than that using a Q-table (4.83 hours)
and gives comparable success rates: 54.8\% for the DQN agent vs.
55.79\% for the agent using Q-tables. The average time to run a single
episode is between 29 and 30 seconds.  
Fig.~\ref{fig:training_case_tb} shows
2 episodes from the same initial position for the ego and ado
vehicles where the ado is able to induce a collision by the ego. 
\begin{figure}[t]
\centering
\subfloat[Adversarial vehicle changes lane far away from the ego vehicle.]{
\includegraphics[width=0.3\linewidth]{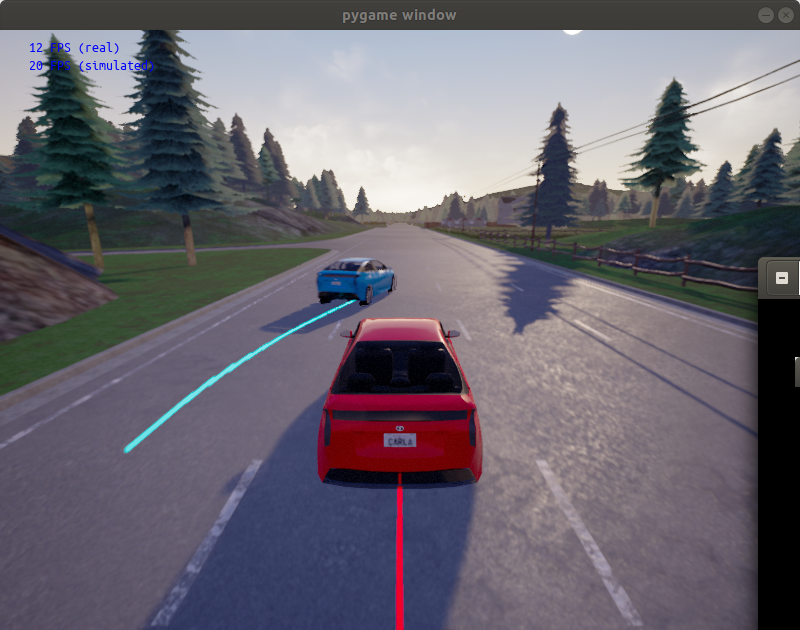}\label{chane_lane_traj_1}}
\hfill
\subfloat[The adversarial vehicle changes lane aggressively and hits the ego vehicle, violating traffic rules.]{
\includegraphics[width=0.3\linewidth]{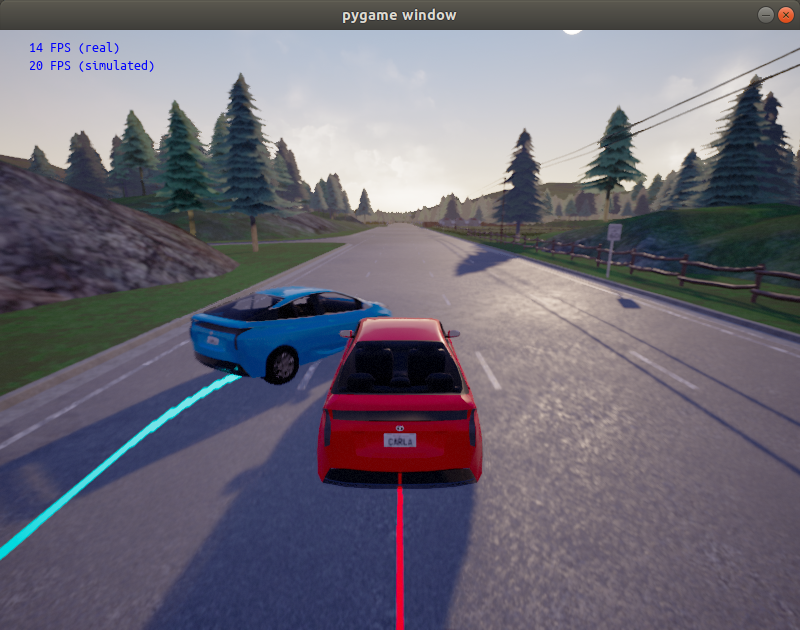}\label{change_lane_traj_5}}
\hfill
\subfloat[Adversary changes lane and induces a crash without braking the traffic rules]{
\includegraphics[width=0.3\linewidth]{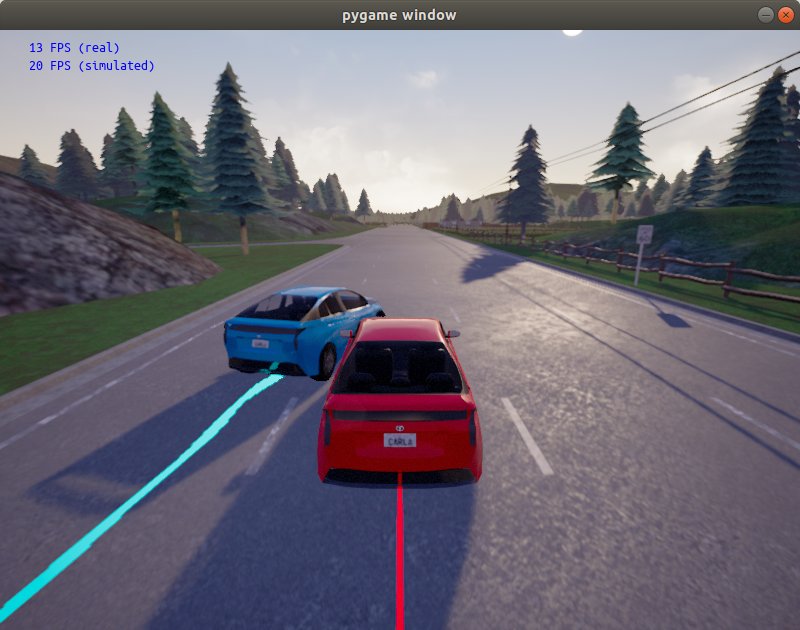}\label{change_lane_traj_6}}
\caption{Ado behaviors across episodes in the lane change maneuvers case study.}\label{fig_case_2}
\end{figure}

\mypara{Lane Change Maneuvers} Here, 2 vehicles are
driving on a two-lane highway;  the ego controlled by a
switching controller that alternates between cruising and
avoiding a collision by applying a ``hard'' brake.  The ego
controller predicts future ado positions based on the current
state using a look-ahead distance $\dlka = \distlat - \vlatado t_{\mathrm{lka}}$,
where $\distlat$ is the lateral distance between the vehicles,
$\vlatado$ is the ado's lateral velocity and $t_{\mathrm{lka}}$ is a fixed
look-ahead time. Based on $\dlka$, it switches between two control
policies: if $\dlka > \dsafe$, then it continues to cruise, but if
$\dlka \le \dsafe$, it applies brakes.  The ado is in
the left lane and attempts to merge to the right in a way that causes
the ego to collide with it. 
We add a constraint to ensure that the ado
should always be longitudinally in front of the ego car when it tries to merge as specified
in Eq.~\eqref{eq:lcadorule}; here $\distlong$ is the longitudinal
distance between the cars. Without this constraint, the ado can always
induce a sideways-crash.
The ego spec is given in
Eq.~\eqref{eq:lcegospec}. Here, $\disttwo = \sqrt{\distlong^2 +
\distlat^2}$ is the Euclidean distance between the two cars.

In the course of training, we observe that the behavior
of the ado agent improves with time. Training for 106 episodes
requires 2.53 hours and gives us a success rate of $71\%$, i.e.
$71\%$ of the episodes lead to a collision induced in the ego. With
206 episodes, the success rate improves to $72.35\%$ but
requires 4.71 hours of runtime. After 371 episodes, the
rate improves to $75.76\%$ after 9.44 hours. This experiment
demonstrates that even under a small time budget, the constrained RL
algorithm can achieve a high amount of success. 

\myipara{Generalizability}
In both case studies, we observed generalizability of the ado policy to different initial
conditions. In the ACC case study, we found several initial states within $\delta=6.5\times
10^{-6}$ that were also counterexample states. Overall the states have smaller values
as the episode lengths are longer and the $\gamma^T$ term causes values to be small. We 
observed that for the failing initial state $(\vego \mapsto 12, \vado \mapsto 12, \dist
\mapsto 15)$, we found failing initial states with values of both $\vado$ and $\dist$ 
that were both smaller and larger than those in the original initial state. However, some
failing initial states did not have states with nearby values that were violating. This
can be attributed to the fact that the RL algorithm may not have converged to a value close
to optimal.  For the second case study, we found that the failing initial condition
$(\vego \mapsto 12,\vado \mapsto 12, \disttwo \mapsto 16)$ has several nearby failing
initial states with a small value of $\delta$ that were not previously encountered
during training.

\section{Related Work and Conclusions }
\label{sec:related_conc}
\myipara{Adaptive Stress Testing} The work of \cite{koren2018adaptive,corso2019adaptive}
is closely related to our work. In this work, the authors also use deep RL
(and related Monte Carlo Tree search) algorithms to seek behaviors
of the vehicle under test that are failure scenarios. There a few
key differences in our approach. In \cite{koren2018adaptive}, reward functions
(that encode failure scenarios) are hand-crafted and require manual
insight to make sure that the RL algorithms converge to behaviors
that are failure scenarios.  Furthermore, the constraints on the
adversarial environment are also explicitly specified. The approach
in\cite{corso2019adaptive} uses a subset of RSS
(Responsibility-Sensitive Safety) rules that are used to augment
hand-crafted rewards to encode failure
scenarios by the ego and responsible behavior by other agents in a
scenario. In specifying STL constraints, we remove the step of
manually crafting rewards.  

\myipara{Falsification} There is extensive related work in
falsification of cyber-physical system. Most falsification techniques
use fixed finitary parameterization of system input signals to define
a finite-dimensional search space, and use global optimizers to search
for parameter values that lead to violation of the system
specification. A detailed survey of falsification techniques can be
found in \cite{deshmukh2019formal}. A control-theoretic view of
falsification tools is that they learn open-loop adversarial policies
for falsifying a given ego model while our approach focuses on
closed-loop policies.

\myipara{Falsification using RL} Also close to our work are recent
approaches to use RL \cite{10.1007/978-3-030-25540-4_23} and deep RL
\cite{akazaki2018falsification} for falsification.  The key focus in
\cite{10.1007/978-3-030-25540-4_23} is on solving the problem of
automatically scaling quantitative semantics for predicates and
effective handling of Boolean connectives in an STL formula. The work
in \cite{akazaki2018falsification} focuses on a smooth approximation
of the robustness of STL and thoroughly benchmarks the use of
different deep RL solvers for falsification.

\myipara{Comparison} Compared to previous approaches, the focus of
our paper is on reusability of dynamically constrained
adversarial agents trained using RL techniques. We identify conditions
under which a trained adversarial policy is applicable to a system
with a different initial condition or different dynamics {\em with no
retraining}. This can be of immense value in an incremental design and
verification approach. The other main contribution is that instead of
using a monolithic falsifier, our technique packs  \emph{multiple,
dynamically constrained} falsification engines as separate agents;
dynamic constraints allow us to specify hierarchical traffic rules.
Furthermore,  previous approaches for falsification do not consider
dynamic constraints on the environment at all -- constraints are
limited to simple bounds on the parameter space. Finally, in our
approach, both specifications and constraints are combined into a
single reward function which can then utilize off-the-shelf deep RL
algorithms.  In comparison to
\cite{10.1007/978-3-030-25540-4_23,akazaki2018falsification,leung2019backpropagation}, our
encoding of STL formulas into reward functions is simplistic as it is
not the main focus of this paper; we defer extensions that consider
nuanced encoding of STL constraints to future work.



\mypara{Conclusions} Our work addresses the problem of automatically
performing constrained stress-testing of cyberphysical systems. We use
STL to specify the target against which we are testing and constraints
that specify reasonableness of the testing regime. We are using STL as
a lightweight, high-level  programming language to loosely specify the
desired behaviors of a test scenario, and leveraging powerful RL
algorithms to determine how to execute those behaviors. The learned
adversarial policies are reactive, as opposed to testing schemes that
rely on merely replaying pre-recorded behaviors, and under limited
conditions can even provide valuable testing capability to modified
versions of the system.


\bibliographystyle{IEEEtran}
\bibliography{bibliography,stl_bibliography,cav_2021_supplement,FMCAD_2021}

\clearpage
\newpage
\section*{Appendix}
\begin{figure}[h!]
    \centering
    \includegraphics[width=\linewidth]{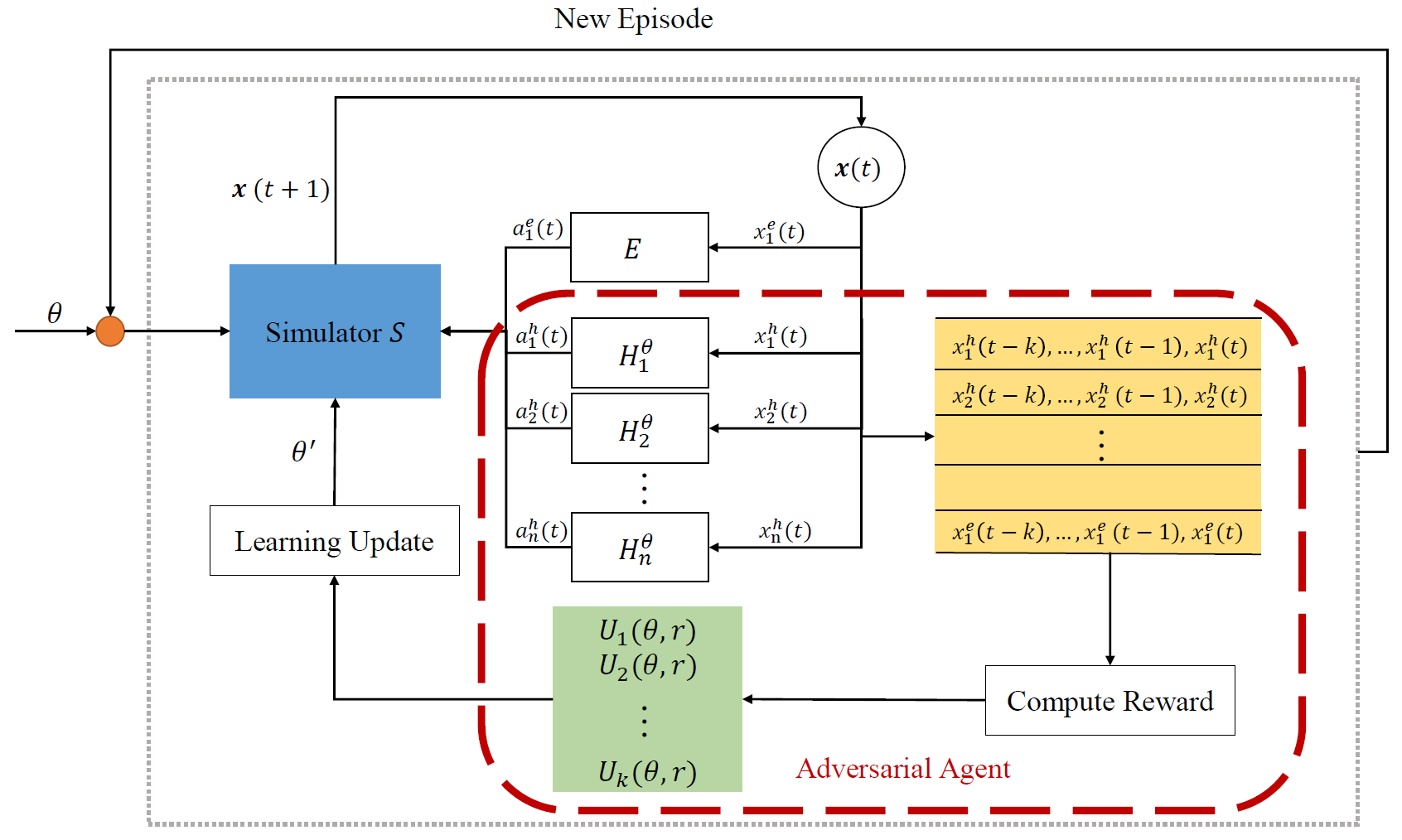}
    \caption{The ego agent $E$ is embedded in a simulation with a
             collection of adversarial agents $H_i^\theta$,
             which learn (possibly from a bank of past experience) 
             to stress-test the ego by a particular reward function
             as derived from the rulebook constraints for the adversary
             and the ego specification.}
    \label{fig:pipeline}
\end{figure}

\mypara{Tool Pipeline}
The tool pipeline is illustrated in Figure \ref{fig:pipeline}. 
The ego agent interacts with several ado agents as part of a simulation. The ados are able to observe the state of the ego as well as of the other ado agents, and they may update their policies. 

\begin{lemma}[Soundness of the reward function]\label{lem:reward_soundness1}
Consider two traces, \(\traj_1\) and \(\traj_2\). 
Suppose that the highest priority constraint violated by 
\(\traj_1\)
is
\(c_1\) and the highest priority constraint violated by
\(\traj_2\)
is \(c_2\). Suppose
\(c_1\) has lower priority than \(c_2\). Then,
the reward for trajectory 1 will be higher than the reward for trajectory 2, i.e.
\(\reward(\traj_1) > \reward(\traj_2)\).
\end{lemma}
\begin{proof}
Let \(n_1\) be the number of rules with priority less than or equal to \(c_1\). Similarly, 
let \(n_2\) the number of rules with priority less than or equal to \(c_2\). Then,
\(\reward(\traj_1) = - n_1 \rho_{min}\) and
\(\reward(\traj_2) = - n_2 \rho_{min}\).
Since \(n_2 > n_1\) by the assumptions of the lemma, the result follows.
\end{proof}

\mypara{Yellow Light} In this experiment, the ego vehicle is
approaching a yellow traffic light led by an ado vehicle.  Let the
signed distance of the ego, ado from the light be respectively
$\dlego$, $\dlado$, and Boolean variables $\elly$ and $\ellr$ be true
if the light is respectively yellow and red.  We use the convention
that $\dlego > -\delta$ if the ego vehicle is approaching the light and
$\dlego < -\delta$ if it has passed the light (resp. for ado).  By traffic
rules, a vehicle is expected to stop $\delta$ meters away from the
traffic light (e.g. $\delta$ could be the width of the intersection
being controlled by the light), I.e. the vehicle should stop at $0$.  Thus,
if $\dlego \in [-\delta,0]$ when the light turns red, it 
has run the red light. This ego specification is shown in
Eq.~\eqref{eq:ylegospec}.  The traffic light is modeled as a
non-adversarial agent, it merely changes its state based on a
pre-determined schedule.  The goal of the ado vehicle is to make the ego
vehicle run the red light.  The rulebook constraints on the ado vehicle are
that it may not drive backwards and it may not run the red light (shown in
Eqs.~\eqref{eq:yladoruleone},\eqref{eq:yladoruletwo} resp.).  The state of 
the ado agent includes the speed of both vehicles, and relative distance 
between the vehicles.  At the start of an episode, $\dlado = 30$, and 
$\elly$ is true, and $\ellr$ becomes true after \(\tau = 2\) seconds.

The ego controller is a switched mode controller that either uses an
ACC controller or applies the maximum available deceleration
$a_{\ego,\max}$.  At time $t<\tau$, let $d(t)$ denote the distance
required for the ego vehicle to come to a stop before the light turns
red by applying $a_{\ego,\max}$. We can calculate $d(t)$ as
$\vego\cdot (\tau-t) + 0.5\cdot a_{\ego,\max} \cdot (\tau-t)^2$. Then,
at time $t$, the ego controller chooses to cruise if $d(t) + \delta <
\dlego$, and brakes otherwise.

\begin{figure}[h!]
    \centering
    \includegraphics[height=5cm]{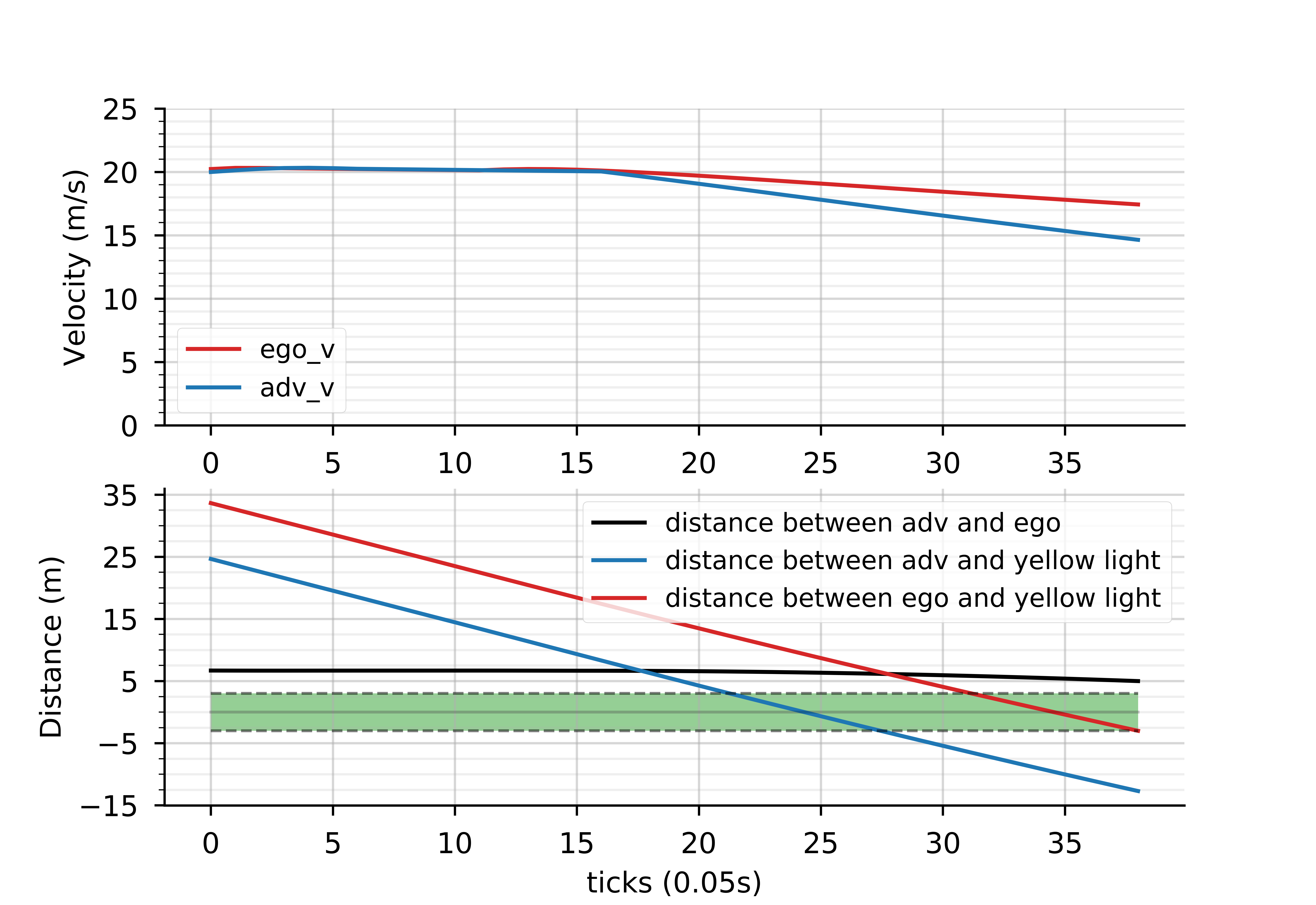}
    \caption{Yellow light case study. The green region represents the region in which the ego vehicle will run the yellow light. The adversary learns to drive the ego car into the target region.}
    \label{fig:yellow_light}
\end{figure}

Figure \ref{fig:yellow_light} shows that the ego vehicle maintains an
appropriate distance to the lead car, but that it starts decelerating
too late and is thus caught in the intersection when light turns red.
The adversarial vehicle successfully clears the intersection while the
light is still yellow, consistent with its rulebook constraints. The
ado agent we train uses the DQN RL algorithm. After 162 episodes of
training, approximately 60\% of its episodes find a violation of the
ego specification with a runtime of 1.88 hours. After 247 episodes,
the success rate increases to 64\% with a runtime of 2.59 hours. This
case study demonstrates that our adversarial testing procedure
succeeds even in the presence of multiple ado rules and an interesting
ego specification.

{\footnotesize 
\begin{table}[t]
\begingroup\setlength{\fboxsep}{0pt}
\begin{tabular}{llML}
\toprule
Case Study & Description & \muc{2}{STL Formula}  \\

\midrule

Yellow & Ego: Don't run red light & 
\neg \ev_{[0,T]} (\ellr \wedge \dlego \in [-\delta,0]) & eq:ylegospec
\\
Light  & Ado: Speed Limits &
\alw_{[0,T]} (\vado < v_{\lim}) & eq:yladoruleone \\
       & Ado: Don't run red light &
\neg \ev_{[0,T]} (\ellr \wedge \dlado \in [-\delta,0]) &
eq:yladoruletwo \\
\bottomrule
\end{tabular}
\endgroup
\caption{Ego Specifications and Ado Rules for case studies\label{tab:specsandrules}}
\end{table}}

\end{document}